\pdfoutput=1
\documentclass[11pt]{article}

\usepackage[final]{acl}

\usepackage{times}
\usepackage{latexsym}

\usepackage[T1]{fontenc}
\usepackage[utf8]{inputenc}

\usepackage{microtype}

\usepackage{inconsolata}

\usepackage{graphicx}

\usepackage{hyperref}       % hyperlinks
\usepackage{url}            % simple URL typesetting
\usepackage{booktabs}       % professional-quality tables
\usepackage{amsfonts}       % blackboard math symbols
\usepackage{nicefrac}       % compact symbols for 1/2, etc.
\usepackage{microtype}      % microtypography
\usepackage{xcolor}         % colors
\usepackage{amsmath}
\usepackage{amssymb}
\usepackage{mathtools}
\usepackage{amsthm}
\usepackage{bm}
\usepackage{multirow}
\usepackage{wrapfig}

\usepackage{subcaption}
\usepackage{algorithm}
\usepackage[noend]{algpseudocode}
\usepackage{bbding}

\newcommand{\methodname}{{\sc C-reweight}}

\theoremstyle{plain}
\newtheorem{theorem}{Theorem}[section]

\theoremstyle{definition}
\newtheorem{definition}[theorem]{Definition}

\theoremstyle{remark}

\newcommand{\x}{\bm{x}}

\renewcommand{\P}{\mathcal{P}}

\newcommand{\cA}{\mathcal{A}}

\newcommand{\cV}{\mathcal{V}}

\newcommand{\cP}{\mathcal{P}}

\title{A Watermark for Auto-Regressive Image Generation Models}

\author{
 \textbf{Yihan Wu$^*$},
 \textbf{Xuehao Cui$^*$},
 \textbf{Ruibo Chen},
 \textbf{Georgios Milis},
 \textbf{Heng Huang}\\
 University of Maryland, College Park\\
 \texttt{ywu42@umd.edu}
 }
\begin{document}
\maketitle
\renewcommand{\thefootnote}{\fnsymbol{footnote}}
\footnotetext[1]{Equal contribution.}
\renewcommand{\thefootnote}{\arabic{footnote}}
\begin{abstract}
The rapid evolution of image generation models has revolutionized visual content creation, enabling the synthesis of highly realistic and contextually accurate images for diverse applications. However, the potential for misuse, such as deepfake generation, image-based phishing attacks, and fabrication of misleading visual evidence, underscores the need for robust authenticity verification mechanisms. While traditional statistical watermarking techniques have proven effective for autoregressive language models, their direct adaptation to image generation models encounters significant challenges due to a phenomenon we term retokenization mismatch—a disparity between original and re-tokenized sequences during the image generation process. To overcome this limitation, we propose \methodname, a novel, distortion-free watermarking method explicitly designed for image generation models. By leveraging a clustering-based strategy that treats tokens within the same cluster equivalently, \methodname\ mitigates retokenization mismatch while preserving image fidelity. Extensive evaluations on leading image generation platforms reveal that \methodname\ not only maintains the visual quality of generated images but also improves detectability over existing distortion-free watermarking techniques, setting a new standard for secure and trustworthy image synthesis.
\end{abstract}

\section{Introduction}
Auto-regressive image generation models, which power cutting-edge visual synthesis applications, have made remarkable strides in producing highly realistic and context-aware images. As these models are increasingly integrated into diverse platforms—from content creation and digital art to real-time visual augmentation—their potential for misuse also grows. Malicious actors can exploit these technologies to fabricate convincing visual evidence, generate deepfake imagery for misinformation, or automate phishing schemes with manipulated visuals. Moreover, the proliferation of synthetic images threatens the authenticity of digital media and raises concerns in legal scenarios where image verification is critical. To mitigate these risks, embedding robust and detectable watermarks in auto-regressive image outputs is crucial, ensuring traceability and accountability while protecting against unauthorized manipulation and misuse.

Statistical watermarking techniques have emerged as a promising solution for identifying machine-generated content from autoregressive language models \citep{kirchenbauer2023watermark}. However, extending these methods directly to autoregressive image generation models faces a critical challenge known as retokenization mismatch. Unlike their language-based counterparts, autoregressive image models involve an additional encoding and decoding phase. During image synthesis, the initial image prompt is first transformed into a sequence of discrete tokens through an encoder. This token sequence is then processed using next-token prediction to generate an output sequence, which is finally decoded back into image form. To verify the presence of a watermark, the generated image must be re-encoded into its tokenized representation. However, the re-encoded token sequence often deviates from the original one used for generation, resulting in retokenization mismatch. This inconsistency severely hampers the effectiveness of traditional watermarking methods when applied directly to image models.

To overcome this limitation, we propose \methodname, a novel, distortion-free watermarking technique specifically crafted for autoregressive image generation models. Our key insight is that mismatched token pairs between the original and re-encoded sequences display a higher degree of similarity compared to random token pairs. Capitalizing on this observation, we design a clustering-based watermarking framework that treats tokens within the same cluster as equivalent, effectively mitigating the impact of retokenization mismatch.

We summarize our contributions as follows:
\begin{itemize}
\item We propose \methodname, a novel distortion-free watermarking framework explicitly designed to address the unique challenges posed by autoregressive image generation models. Unlike traditional statistical watermarking techniques, \methodname\ effectively mitigates the impact of retokenization mismatch through a clustering-based strategy, ensuring consistent watermark detectability even after image decoding and re-encoding.

\item We introduce a clustering-based equivalence mechanism that leverages token similarity to enhance watermark resilience. By grouping tokens with high similarity into clusters, our method preserves watermark signals across retokenization processes without altering image fidelity, enabling reliable verification.

\item We perform comprehensive evaluations on state-of-the-art autoregressive image generation models, demonstrating that \methodname\ consistently outperforms existing watermarking methods. Experimental results reveal a 10\% increase in detection accuracy, establishing a new benchmark for robustness and reliability in watermarking synthetic images.

% \item We validate the robustness of \methodname\ under various perturbations. Our method maintains high detectability rates while preserving the visual quality of the watermarked images, illustrating its applicability for real-world deployment.
\end{itemize}

\section{Related Work}
\paragraph{Auto-regressive generative models.} 
Auto-regressive models have played a pivotal role in advancing language modeling, powering state-of-the-art systems such as GPT-3 and ChatGPT~\citep{vaswani2017attention,raffel2020exploring,brown2020language}, and prompting discourse on the emergence of artificial general intelligence. In multimodal domains, however, image generation has been predominantly driven by diffusion-based approaches (e.g., Stable Diffusion~\citep{rombach2022high}), while vision-language understanding has relied on compositional frameworks like CLIP~\citep{radford2021learning} paired with LLMs (e.g., LLaVA~\citep{liu2023visual}). Attempts to unify generation and perception, such as Emu~\citep{sun2023emu} and Chameleon~\citep{team2024chameleon}, either depend on coupling LLMs with diffusion models or fall short of matching the task-specific performance of specialized methods. More recently, Emu3~\citep{wang2024emu3} demonstrates that a unified auto-regressive model trained purely via next-token prediction can achieve state-of-the-art results across vision, language, and video tasks. By tokenizing all modalities into a shared discrete space, Emu3 trains a single transformer from scratch on a diverse mix of multimodal sequences, bypassing the need for diffusion processes or compositional fusion techniques.

% \paragraph{Auto-regressive generative models.}
% ===============================================
% Auto-regressive generative models has revolutionized the field of language models \citep{vaswani2017attention,raffel2020exploring,brown2020language}, enabling breakthroughs like ChatGPT and sparking discussions about the early signs of artificial general intelligence (AGI). In the realm of multimodal models, vision generation has been dominated by complex diffusion models (e.g., Stable Diffusion~\citep{rombach2022high}), while vision-language perception has been led by compositional approaches such as CLIP \citep{radford2021learning} with LLMs (e.g., LLaVA~\citep{liu2023visual}). Despite early attempts at unifying generation and perception, such as Emu~\citep{sun2023emu} and Chameleon~\citep{team2024chameleon}, these efforts either resort to connecting LLMs with diffusion models or fail to match the performance of task-specific methods tailored for generation and perception.
% Emu3~\citep{wang2024emu3}, a new set of state-of-the-art multimodal models based solely on next-token prediction, eliminating the need for diffusion or compositional approaches entirely. Images, text, and videos are tokenized into a discrete space, and jointly train a single transformer from scratch on a mix of multimodal sequences.
% ===============================================

\paragraph{Diffusion model watermarking.} 
Diffusion model watermarking embeds watermarks directly during the image generation process to ensure minimal perceptual distortion. Some techniques achieve this by modifying the generative model itself, for instance, by fine-tuning specific components, as seen in Stable Signature~\citep{fernandez2023stable}. Alternatively, other methods embed watermarks by perturbing the initial noise input, avoiding the need for extensive model retraining. Tree-Ring~\citep{wen2023tree}, for example, inserts a Fourier-domain pattern into the initial noise, which can be recovered via DDIM inversion~\citep{song2020denoising}. RingID~\citep{ci2024ringid} extends this approach to support multiple watermark keys. Additional notable techniques include Gaussian Shading~\citep{yang2024gaussian}, which generates a unique key per watermark owner, and PRC~\citep{gunn2024undetectable}, which employs pseudo-random error-correcting codes to achieve computationally undetectable watermarking. These approaches are orthogonal to our work, as they are specifically tailored for diffusion models and cannot be readily applied to other generative architectures.

\paragraph{Distortion-free watermark.} \cite{Aaronson2022} introduced a pioneering distortion-free watermarking method that leverages Gumbel-Softmax to adjust token distributions without altering content quality. Building on this concept, \cite{christ2023undetectable} and \cite{kuditipudi2023robust} applied inverse-sampling and Gumbel-Softmax techniques, respectively, to modify token distributions in watermarked outputs. Their approaches rely on watermark keys derived from either token positions or predefined key lists. However, \cite{christ2023undetectable}'s method demonstrates limited robustness against modifications and lacks empirical validation of its detectability. In contrast, \cite{kuditipudi2023robust}'s approach necessitates substantial resampling from the secret key distribution for effective detection, resulting in inefficiencies for long-form content. To address these limitations, \cite{hu2023unbiased} introduced inverse-sampling and reweight strategies, although their detection mechanism is not model-agnostic and depends on access to the language model API and specific prompts. Subsequently, \cite{wu2023dipmark} enhanced the reweighting approach and proposed a model-agnostic detection method. Most recently, \cite{dathathri2024scalable} introduced SynthID, achieving distortion-free watermarking for LMs across multiple generations, advancing scalability and detection robustness.
\begin{table*}[h]
    \centering
    \caption{Visual Comparison of generation results with and without \methodname.}

    \includegraphics[width=\textwidth]{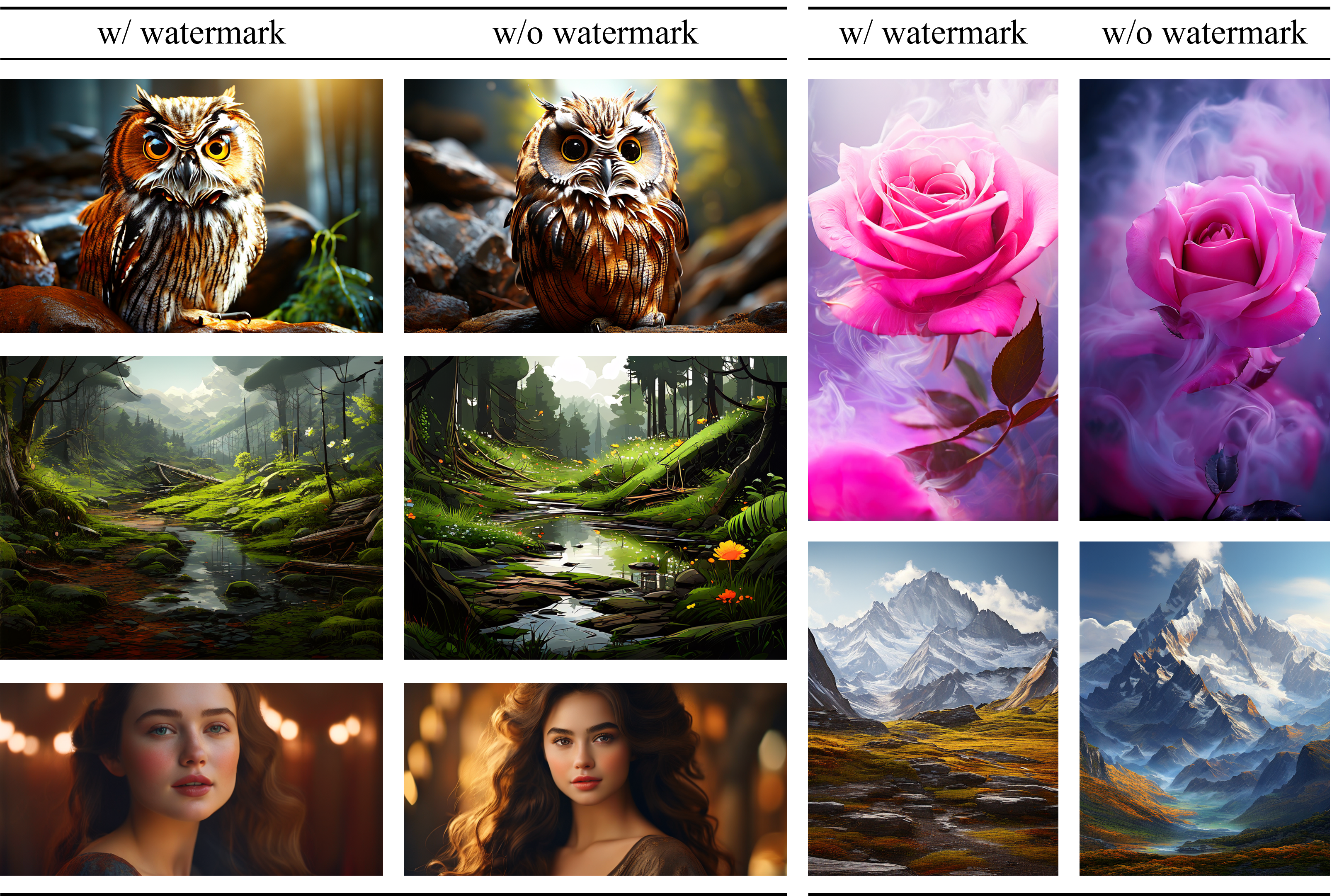}
    \label{tab:main_imqual}
    \vspace{-0.5cm}
\end{table*}
\section{Preliminary}

\paragraph{Notations.} We follow the notations used in \citep{hu2023unbiased}. The vocabulary (or token) set is denoted by $V$ and its cardinality by $N=|V|$. We define the set $\cV$, which includes all possible token sequences including those of zero length and the set $\cA$, which includes all possible images. Within an autoregressive image generation model, a token sequence is generated based on a specific prompt. At any given step, the probability of producing the next token $x_{n+1} \in V$, given the preceding sequence $x_1, \ldots, x_n$, is denoted by $P_M(x_{n+1} \mid x_1, x_2, \ldots, x_n)$.
For simplicity and clarity, we adopt a more concise notation: $P_{M}(\bm{x}_{n+1:n+m} \mid \bm{x}_{1:n})$, where $\bm{x}_{n+1:n+m} = (x_{n+1}, \ldots, x_{n+m})$. It is important to note that the prompt is intentionally excluded from this notation. In image generation models, we denote the image-token encoder by $E(\cdot):\cA\to\cV$ and the token-image decoder by $D(\cdot):\cV\to\cA$.
\subsection{Statistical Watermarks}
In watermarking applications, the service provider employs a set of \textit{i.i.d.} watermark codes $\{\theta_i \in \Theta, i \in \mathbb{N}\}$, defined over the code space $\Theta$. Each code $\theta_i$ is typically derived from a secret key $\textsf{key} \in \mathcal{K}$ and the n-gram preceding context, denoted $\x_{t-n:t-1}$.

In the watermark generator, a reweight strategy is used to embed a statistical signal into the generated content. Let $\mathcal{P}$ denote the set of all probability distributions over the token set $V$. The reweight strategy is a function $P_W : \mathcal{P} \times \Theta \to \mathcal{P}$. For the token distribution at the $(n+1)$-th generation step, $P_M(x_{n+1} \mid \x_{1:n}) \in \mathcal{P}$, the watermarked distribution is defined by $P_W(P_M(x_{n+1} \mid \x_{1:n}), \theta_i)$. For brevity, this is represented as $P_W(x_{n+1} \mid \x_{1:n}, \theta_i)$. A distortion-free watermark ensures that the averaged distribution $P_W(x_{n+1} \mid \x_{1:n}, \theta_i)$ with respect to $\theta_i$ is equal to the original distribution $P_M(x_{n+1} \mid \x_{1:n})$. 
\begin{definition}[Distortion-free watermark]
    Given the watermark code set $\Theta$, a distribution $\P_{\Theta}$ on $\Theta$, original LM distribution $P_M$, and the watermarked distribution $P_W(\cdot|\theta\in\Theta)$ A distortion-free watermark should satisfy $\forall x \in V$,
\[
\mathbb{E}_{\theta \sim P_{\Theta}} \left[P_W(x \mid \x_{1:n}, \theta)\right] = P_{M}(x\mid \x_{1:n}).
\]
\end{definition}

Current popular distortion-free strategies include Gumbel-softmax~\citep{Aaronson2022}, inverse-sampling~\citep{christ2023undetectable, hu2023unbiased, kuditipudi2023robust} and reweight-based strategy~\citep{wu2023dipmark,dathathri2024scalable}.

During watermark detection, the user only has access to the watermark key, the reweight strategy, and the generated image. The detector employs a hypothesis testing approach to ascertain the presence of the watermark signal. The null hypothesis $H_0$ is defined as \textit{``The content is generated without the presence of watermarks"}. The detector adopts a score function based on the watermark key and the reweight strategy, which exhibits statistical bias between the watermarked and unwatermarked token sequences.

% \subsection{Statistical watermarks}
% In watermarking applications, the service provider utilizes a set of \textit{i.i.d.} watermark codes ${\theta_i \in \Theta, i \in \mathbb{N}}$, defined over the code space $\Theta$. Each code $\theta_i$ is typically derived from a secret key $k \in \K$ and a the n-gram preceding context i.e. $\x_{t-n:t-1}$.

% In watermark generator, we use a reweight strategy to embed the statistical signal into the generated content. Let $\P$ denote the set of all probability distributions over the token set $V$. A reweight strategy is a function $P_W : \P \times \Theta \to \P$. For the token distribution at the (n+1)-th generation step $P_M(x_{n+1} \mid \x_{1:n}) \in \P$, the watermarked distribution is defined by $P_W(P_M(x_{n+1} \mid \x_{1:n}), \theta_i)$. For brevity this is represented as $P_W(x_{n+1} \mid \x_{1:n}, \theta_i)$. Current popular distortion-free strategies includes Gumbel-Softmax~\cite{Aaronson2022} and inverse-sampling~\cite{christ2023undetectable,hu2023unbiased,kuditipudi2023robust}.

% During watermark detection, the user only has access to the watermark key, the reweight strategy, and the generated image. The detector uses a hypothesis testing approach to identify the presence of the watermark signal. The null hypothesis $H_0$ is \textit{``The content is generated without the presence of watermarks."} The detector will adopt a score function based on the watermark key and the reweight strategy, which presents statistical bias  between the watermarked and unwatermarked tokens sequences. 

\subsection{Autoregressive image generation models}
% ===============================================
The vision tokenizer of Emu3 is trained based on SBER-MoVQGAN~\citep{zheng2022movq}, which can encode a 4 × 512 × 512 video clip or a 512 × 512 image into 4096 discrete tokens from a codebook of size 32,768. The tokenizer achieves 4× compression in the temporal dimension and 8×8 compression in the spatial dimension, applicable to any temporal and spatial resolution. Building on the MoVQGAN architecture~\citep{zheng2022movq}, we incorporate two temporal residual layers with 3D convolution kernels into both the encoder and decoder modules to enhance video tokenization capabilities. The tokenizer is trained end-to-end on the LAION-High-Resolution image dataset~\citep{schuhmann2022laion} and the InternVid~\citep{wang2023internvid} video dataset using combined objective functions of L2 loss, LPIPS perceptual loss~\citep{zhang2018unreasonable}, GAN loss, and commitment loss~\citep{esser2021taming}.

\paragraph{Retokenization mismatch.} Let $\x$ represent the output token sequence of the image generation model, where the output image is decoded through the image-token decoder $D(\x)$. During watermark detection, the generated image is re-encoded through the encoder $E(D(\x))$, and the watermark signal within $E(D(\x))$ is detected via a hypothesis test. However, there is typically a discrepancy between $E(D(\x))$ and $\x$, which can diminish the statistical signal introduced by the detection scores. We can interpret this as an inevitable token-level edit attack during detection.

% \paragraph{retokenization mismatch.} Denoted by $\x$ the output token sequence of the image generation model, the output image is decoded though the image-token decoder $D(\x)$. During the watermark detection, we will re-encode the generated image through the encoder $E(D(\x))$, and detect the watermark signal within $E(D(\x))$ via a hypothesis test. However, there is usually a gap between $E(D(\x))$ and $\x$, which will weaken the statistical signal introduced by the detector scores. 
\begin{algorithm}[h]
\caption{Cluster-based reweight.}\label{alg:aligned}
\begin{algorithmic}[1]
\State \textbf{Input:} Original model distribution $P_M(\cdot|\x_{1:n})$. Watermark code $\theta$.
% \State Initialize the simi
% \State Initialize a list to store the overlapped regions: overlapped_vec=[]
\State Calculate cluster probabilities $\Pr(c_i):=\sum_{x\in c_{i}}P_M(x|\x_{1:n}),i\in[h]$.
\State Calculate distribution $\cP_{c}:=norm(\min\{0,h\Pr(c_{i})-1\},...,\min\{0,h\Pr(c_{i})-1\})$.
\State Pseudo-randomly select a cluster $c_{i'(\theta)}$.
\State Randomly select $j\sim Uniform([0,1])$.
\If{$j<\Pr(c_{i'(\theta)})$}
\State $c_{i(\theta)}=c_{i'(\theta)}$
\Else 
\State Sampling $c_{i''(\theta)}\sim\cP_{c}$.
\State $c_{i(\theta)}=c_{i''(\theta)}$
\EndIf
\State Sampling next token $x$ via Eq.~\ref{eqn:second step sampling}.
\State \textbf{return} $x$.
\end{algorithmic}
% \vspace{-0.5cm}
\end{algorithm}

\section{Methodology}
To mitigate retokenization mismatches in autoregressive image generators, we propose \methodname, a clustering-based watermarking framework. First, the discrete image tokens are grouped into clusters. On top of these clusters we design a distortion-free, cluster-specific reweighting rule that perturbs the model’s output probabilities only at the cluster level. During generation, this rule steers sampling so that each emitted token lies in the cluster dictated by the secret watermark code; during detection, we simply test whether the recovered token belongs to the prescribed cluster. Because a retokenization error usually maps a token to another member of the same cluster, the detector still receives a reliable statistical signal.

The remainder of this section is organized as follows. We begin by describing the clustering procedure and the accompanying cluster-based reweight strategy that preserves distortion while counteracting retokenization errors. We then present the generic watermarking algorithm built on this strategy, together with its associated detection statistic.

\subsection{Cluster-based reweight strategy}

\paragraph{Image token clustering.} In image token clustering, the objective is to split all image tokens into distinct clusters based on their similarity. It is crucial to note that the clustering algorithm is executed only once for each image generation model. After the clusters have been established, there is no need to rerun the algorithm during watermark generation and detection. Consequently, this approach does not significantly increase computational costs. To achieve this, we collect the image token embeddings $\{e_1,...,e_m\}$ based on the token-image decoder $D$ and use k-means algorithm to generate the corresponding clusters $\{c_1,...,c_h\}$.
With the above-mentioned segmentation strategy, mismatched tokens are more likely to be in the same cluster because they are supposed to share similar embeddings in the decoder $D$. The watermark generator and detector can utilize the cluster information to avoid the detectability reduction caused by the retokenization mismatch. The inverse sampling watermark is a distortion-free method that can be applied directly to clustering scenarios.

\paragraph{DiP-reweight.} DiP-reweight~\citep{wu2023dipmark} is a representative example of a two-cluster distortion-free reweighting strategy. In DiP-reweight, the token set is partitioned into two clusters: a red list and a green list. The method aims to promote the total probability mass of the green tokens. Specifically, token probabilities are arranged along the interval $[0,1]$, with red tokens placed at the beginning and green tokens at the end. To achieve distortion-free watermarking, a parameter $\alpha \leq 0.5$ is selected such that probabilities in $[0, \alpha]$ are removed, those in $[\alpha, 1-\alpha]$ remain unchanged, and those in $[1-\alpha, 1]$ are doubled. However, DiP-reweight is restricted to two-cluster settings, limiting its applicability in scenarios involving multiple clusters. To overcome this limitation, we propose a novel cluster-based reweighting strategy tailored for multi-cluster settings.

\paragraph{Cluster-based reweight.} Denote by the token clusters $\{c_1,...,c_h\}$. Given the preceding tokens $\x_{1:n}$, the cluster-based reweight consists of two steps. In the first step, we pseudo-randomly select a token cluster $c_{i(\theta)}$ (with equal probability) based on the watermark code $\theta$ through reject sampling. In the second step, we randomly sample a token from $c_{i(\theta)}$ based on the original model's distribution.

For the first step, we will first calculate the cluster probabilities $\Pr(c_1),...,\Pr(c_h)$ via $\Pr(c_i):=\sum_{x\in c_{i}}P_M(x|\x_{1:n})$. We will first pseudo-randomly selected a cluster $c_{i'(\theta)}$ based on the watermark code $\theta$. Then we randomly select a number $j\in[0,1]$, if $j<h\Pr(c_{i'(\theta)})$, we will sample a token within $c_{i'(\theta)}$. Otherwise, we reject this sample and sample another cluster $c_{i''(\theta)}$ based on the overflow probability distribution $\cP_{c}:=norm(\min\{0,h\Pr(c_{i})-1\},...,\min\{0,h\Pr(c_{i})-1\})$, where $norm$ is a distribution normalization function.

Given the selected cluster $c_{i(\theta)}$, in the second step we randomly sample a token from $c_{i(\theta)}$ based on the original model's token probability, i.e., 
\begin{equation}\label{eqn:second step sampling}
        P_W(x|\x_{1:n},\theta) := \left\{\begin{array}{cl}
            \frac{P_M(x|\x_{1:n})}{\Pr(c_{i(\theta)})},& \text{if } x\in c_{i(\theta)} \\
            0, & \text{if } x\notin c_{i(\theta)}.
        \end{array}\right.
    \end{equation}
Details on this method are in Algorithm~\ref{alg:aligned}.

\begin{theorem}
    Cluster-based reweight is distortion-free.
\end{theorem}
The detailed proof can be found in Appendix~\ref{sec:missing proof}

\begin{algorithm}[t]
\caption{\methodname\ generator.}\label{alg:generator}
\begin{algorithmic}[1]
\State \textbf{Input:} secret key $\textsf{key}$, prompt $\bm{x}_{-m:0}$, generate length $t\in\mathbb{N}$, token-image decoder $D$.
% \State \textbf{Input:}
% secret key $\sk$, parameter $\beta$, prompt $\bm{x}_{-m:0}$, generate length $n\in\mathbb{N}$, texture key history $hist$, n-gram parameter $a$, and permutation generation function $h$.
\State Initialize watermark code history $hist$.
\For{$i=1,\dots,t$}
        \State Calculate the token distribution for generating the $i$-th token $P_M(\cdot\mid\bm{x}_{-m:i-1})$. %\Comment{original distribution}
        \State Generate a watermark code $\theta_{i} = ($\textsf{key}$,\x_{i-n,i-1})$.
        \If{$\theta_i\in hist$}
        \State Sample the next token $x_{i}$ using distribution 
        \Else
        \State Generate the pseudo-random number $r(\theta_i)$.
        \State Calculate watermarked distribution $P_W(\cdot|\x_{-m:i-1})$ via cluster-based reweight.
    \State Sample the next token $x_{i}$ using distribution $P_W(\cdot|\x_{-m:i-1})$.
    \EndIf
\EndFor
\State \textbf{return} image $D(\bm{x}_{1:t})$.
\end{algorithmic}
% \vspace{-0.5cm}
\end{algorithm}

\begin{algorithm}[t]
\caption{\methodname\ detector.}\label{alg:detector}
\begin{algorithmic}[1]
\State \textbf{Input:} image $a$, image-token encoder $E$, secret key \textsf{key}, score function $s$, threshold $z$.
% \State Initialize the simi
\State calculate token sequence $\x_{1:t}=E(a)$
% \State Initialize a list to store the overlapped regions: overlapped_vec=[]
\State Initialize the score function: $S=0$.
    \For{$i = 2,...,t$}
    \State Generate the watermark code $k_{i} = (\textrm{key},\x_{i-n,i-1})$.
    \State Generate the pseudo-random number $r(\theta_i)$.
    \State Update the score function via $S = S + s(\theta_i,x_i)$.
    \EndFor
\State \textbf{return} $S>z$.
\end{algorithmic}
% \vspace{-0.5cm}
\end{algorithm}
\begin{table*}[h!]
\centering
\caption{\normalsize Detectability comparison with KGW~\citep{kirchenbauer2023watermark}, Unigram~\citep{hu2023unbiased} and DiPmarkk~\citep{wu2023dipmark} on Pickapic\_v2~\citep{kirstain2023pick}, MSCOCO~\citep{lin2014microsoft}, Flickr8k~\citep{hodosh2013framing} and LAION~\citep{schuhmann2022laion} datasets. [\textbf{KEY}: best]}
\label{tab:main_res}
\resizebox{0.9\textwidth}{!}{
\begin{tabular}{l|cc|cc|cc|cc} 
\toprule\normalsize
Dataset & \multicolumn{2}{|c|}{Pickapic\_v2} & \multicolumn{2}{|c|}{MSCOCO} & \multicolumn{2}{|c|}{Flickr8k} & \multicolumn{2}{|c}{LAION} \\
\midrule
 % TPR & @FPR=1\%$\uparrow$ & TPR@FPR=0.1\%$\uparrow$ & TPR@FPR=0.01\%$\uparrow$ & Median p-value$\downarrow$ \\ \midrule
 FPR & 1\% & 0.1\% & 1\% & 0.1\% & 1\% & 0.1\% & 1\% & 0.1\% \\ \midrule\midrule
KGW($\delta$=0.5) & 0.31 & 0.13 & 0.34 & 0.15 & 0.48 & 0.22 & 0.24 & 0.10 \\
KGW($\delta$=1.0) & 0.75 & 0.53 & 0.75 & 0.52 & 0.80 & 0.68 & 0.75 & 0.49 \\
KGW($\delta$=1.5) & 0.92 & 0.77 & 0.92 & 0.82 & 0.96 & 0.94 & 0.89 & 0.79 \\
KGW($\delta$=2.0) & 0.94 & 0.91 & 0.95 & 0.93 & 0.97 & 0.95 & 0.95 & 0.88 \\
Unigram($\delta$=0.5) & 0.79 & 0.59 & 0.82 & 0.65 & 0.85 & 0.78 & 0.83 & 0.65 \\
Unigram($\delta$=1.0) & 0.96 & 0.86 & 0.94 & 0.88 & \textbf{0.99} & 0.97 & 0.93 & 0.86 \\
Unigram($\delta$=1.5) & 0.97 & 0.96 & \textbf{0.99} & 0.98 & \textbf{0.99} & 0.98 & 0.98 & 0.97 \\
Unigram($\delta$=2.0) & \textbf{0.99} & 0.96 & \textbf{0.99} & \textbf{0.99} & \textbf{0.99} & \textbf{0.99} & 0.98 & 0.98 \\ \midrule
ITS-edit & 0.51 & 0.41 & 0.43 & 0.34 & 0.59 & 0.53 & 0.56 & 0.43 \\
EXP-edit & 0.94 & 0.92 & 0.93 & 0.90 & 0.95 & 0.94 & 0.93 & 0.92 \\
$\gamma$-reweight & 0.44 & 0.31 & 0.51 & 0.35 & 0.51 & 0.38 & 0.45 & 0.23 \\
DiPmark($\alpha$=0.4) & 0.24 & 0.12 & 0.36 & 0.16 & 0.52 & 0.34 & 0.25 & 0.09 \\
DiPmark($\alpha$=0.3) & 0.16 & 0.05 & 0.15 & 0.06 & 0.26 & 0.10 & 0.09 & 0.04 \\
STA-1 & 0.24 & 0.09 & 0.25 & 0.12 & 0.20 & 0.07 & 0.21 & 0.07\\
\midrule
\methodname($h$=100) & 0.96 & 0.94 & \textbf{0.99} & \textbf{0.99} & \textbf{0.99} & 0.95 & \textbf{0.99} & \textbf{0.99} \\
\methodname($h$=200) & \textbf{0.99} & \textbf{0.98} & \textbf{0.99} & \textbf{0.99} & \textbf{0.99} & 0.98 & \textbf{0.99} & 0.98 \\ \bottomrule
\end{tabular}
}
\end{table*}

With the aforementioned probability assignments, we can define a statistical score for the detector based on the watermark code $\theta$ and the current token $x$. 
\begin{definition}\label{def:detection score}
    Given the watermark code $\theta$ and the current token $x$, the detection score $s(\theta,x)$ is defined as 
    \begin{equation}
        s(\theta,x) := \left\{\begin{array}{cl}
            1 & \text{if } x \in c_{i'(\theta)}, \\
            0 & \text{otherwise}.
        \end{array}\right.
    \end{equation}
\end{definition}
Note, when $\Pr(c_{i'(\theta)}) < \frac{1}{h}$, cluster-based reweight may sample tokens from clusters other than $c_{i'(\theta)}$. This can reduce detection accuracy. However, empirical results indicate that the overall detectability of cluster-based reweight still surpasses that of the state-of-the-art distortion-free reweight strategy.

\subsection{\methodname}

Leveraging the cluster-based reweight strategy, we design our watermarking scheme, \methodname, which contains two main components: a \emph{watermark generator} and a \emph{watermark detector}.

\paragraph{Watermark generator.}
At generation step~$t$, a watermark code $\theta_t$ is derived from the secret key $\textsf{key}$ and the $n$-gram context $\x_{t-n,t-1}$.
This code first selects a reference cluster $c_{i'(\theta_t)}$.
Following the reweight rule, the actual sampling cluster $c_{i(\theta_t)}$ is chosen using the original cluster probability $\Pr\!\bigl(c_{i'(\theta_t)}\bigr)$ and the overflow distribution $\cP_c$.
The next token $x_t$ is then sampled from $c_{i(\theta_t)}$ according to the base model distribution $P_M(\cdot \mid \x_{1:t-1})$.
To preserve distortion-freeness across multiple generations, we adopt the watermark-code history $hist$ of~\cite{hu2023unbiased}: whenever $\theta_t \in hist$, we revert to sampling directly from the unmodified model distribution.
The full procedure is given in Alg.~\ref{alg:generator}.

\paragraph{Watermark detector.}
Detection assumes access to the generated image and the same key~$\textsf{key}$.
The image encoder $E$ recovers the token sequence $\x_{1:t}$.
For each position $i=1,\dots,t$, we recompute the code $\theta_i$ from the key and local $n$-gram context, then evaluate the detection statistic $s(\theta_i,x_i)$ from Definition~\ref{def:detection score}.
The aggregate statistic is
\[
S(\x_{1:t}) \;=\; \sum_{i=1}^{t} s(\theta_i, x_i),
\]
and Alg.~\ref{alg:detector} lists the full algorithm.

\paragraph{Statistical test.}
Under the null hypothesis (no watermark), $S(\x_{1:t})$ follows a binomial distribution with success probability $\tfrac{1}{h}$, yielding the tail bound
\[
\Pr\!\bigl(S(\x_{1:t}) \ge k\bigr)
    \;=\;
    \sum_{i=\lceil k\rceil}^{t}
        \binom{t}{i}
        \Bigl(\tfrac{1}{h}\Bigr)^{i}
        \Bigl(\tfrac{h-1}{h}\Bigr)^{\,t-i}.
\]

\begin{table}[t]
    \centering
    \caption{Robustness comparison on detectability under $l_2$ noise attacks. [\textbf{KEY}: best]}
    \label{tab:robustness_l2}
    \begin{tabular}{lccc}
        \toprule
        & \multicolumn{3}{c}{$l_2$} \\
        \cmidrule(lr){2-4}
       Budget & 0.25 & 0.50 & 1.00 \\ \midrule
       KGW($\delta$=0.5) & 0.26 & 0.24 & 0.24 \\
       KGW($\delta$=1.0) & 0.60 & 0.65 & 0.69 \\
       KGW($\delta$=1.5) & 0.81 & 0.79 & 0.80 \\
       KGW($\delta$=2.0) & 0.94 & 0.94 & 0.94 \\
       Unigram($\delta$=0.5) & 0.78 & 0.61 & 0.55 \\
       Unigram($\delta$=1.0) & 0.92 & 0.88 & 0.87 \\
       Unigram($\delta$=1.5) & 0.97 & 0.97 & 0.96 \\
       Unigram($\delta$=2.0) & \textbf{0.99} & 0.96 & 0.97 \\ 
       ITS-edit & 0.34 & 0.39 & 0.43 \\
       EXP-edit & 0.88 & 0.89 & 0.88 \\
       $\gamma$-reweight & 0.36 & 0.38 & 0.35 \\
       DiPmark($\alpha$=0.4) & 0.22 & 0.22 & 0.23 \\
       DiPmark($\alpha$=0.3) & 0.12 & 0.13 & 0.08 \\
       STA-1 & 0.21 & 0.20 & 0.22 \\
       \methodname($h$=200) & \textbf{0.99} & \textbf{0.99} & \textbf{0.99} \\
       \bottomrule
    \end{tabular}
\end{table}
\begin{table}[t]
    \centering
    \caption{Robustness comparison on detectability under $l_{\infty}$ noise attacks. [\textbf{KEY}: best]}
    \label{tab:robustness_linf}
    \begin{tabular}{lccc}
        \toprule
        & \multicolumn{3}{c}{$l_{\infty}$} \\
        \cmidrule(lr){2-4}
       Budget & 2/255 & 4/255 & 8/255 \\ \midrule
       KGW($\delta$=0.5) & 0.15 & 0.06 & 0.06 \\
       KGW($\delta$=1.0) & 0.48 & 0.17 & 0.13 \\
       KGW($\delta$=1.5) & 0.64 & 0.35 & 0.14 \\
       KGW($\delta$=2.0) & 0.68 & 0.39 & 0.19 \\
       Unigram($\delta$=0.5) & 0.70 & 0.62 & 0.47 \\
       Unigram($\delta$=1.0) & 0.90 & 0.79 & 0.59 \\
       Unigram($\delta$=1.5) & \textbf{0.97} & 0.84 & 0.72 \\
       Unigram($\delta$=2.0) & 0.95 & 0.90 & \textbf{0.79} \\ 
       ITS-edit & 0.17 & 0.02 & 0.01 \\
       EXP-edit & 0.74 & 0.55 & 0.33 \\
       $\gamma$-reweight & 0.21 & 0.08 & 0.05 \\
       DiPmark($\alpha$=0.4) & 0.12 & 0.05 & 0.01 \\
       DiPmark($\alpha$=0.3) & 0.07 & 0.05 & 0.02 \\
       STA-1 & 0.17 & 0.06 & 0.03 \\
       \methodname($h$=200) & 0.93 & \textbf{0.91} & 0.78 \\
       \bottomrule
    \end{tabular}
\end{table}
\begin{table*}
    \label{table:scannability}
    \centering
    \caption{Quantitative measurement of generated images.}
    \resizebox{\textwidth}{!}{
    \begin{tabular}{lcccccccc}
        \toprule
        
        & \multicolumn{2}{c}{Pickapic\_v2} & \multicolumn{2}{c}{MSCOCO} & \multicolumn{2}{c}{Flickr8k} & \multicolumn{2}{c}{LAION}   \\ \cmidrule(lr){2-3} \cmidrule(lr){4-5} \cmidrule(lr){6-7} \cmidrule(lr){8-9}  
        & FID$\downarrow$ & CLIP$\uparrow$ & FID$\downarrow$ & CLIP$\uparrow$ & FID$\downarrow$ & CLIP$\uparrow$ & FID$\downarrow$ & CLIP$\uparrow$  \\
        \midrule
        w/o watermark & 162.06 &0.854& 136.41 &0.872& 153.18 &0.870& 112.36 &0.906\\ \midrule 
        % KGW($\delta$=0.5) & 151.56 &0.859& 138.40 &0.870& 150.48 &0.860& 122.87 & 0.903\\
        % KGW($\delta$=1.0) & 150.90 &0.858& 139.79 &0.872& 153.55 &0.859& 118.93 & 0.902\\
        % KGW($\delta$=1.5) & 150.48 &0.859& 133.30 &0.868& 146.85 &0.865& 122.05 & 0.900\\
        % KGW($\delta$=2.0) & 152.82 &0.864& 136.51 &0.864& 150.90 &0.870& 118.67 & 0.904\\
        % Unigram($\delta$=0.5) & 151.52 &0.860& 140.59 &0.870& 152.61 &0.860& 119.86 & 0.901\\
        % Unigram($\delta$=1.0) & 156.27 &0.861& 139.33 &0.868& 149.19 &0.869& 120.21 & 0.900\\
        % Unigram($\delta$=1.5) & 153.78 &0.855& 138.74 &0.866& 151.91 &0.866& 118.64 & 0.904\\
        % Unigram($\delta$=2.0) & 155.50 &0.863& 141.64 &0.865& 150.46 &0.860& 119.72 & 0.906\\ \midrule
        ITS-edit & 163.02 &0.857& 136.00 &0.875& 149.76 &0.861& 117.11 & 0.906\\
        EXP-edit & 151.17 &0.850& 140.61 &0.868& 151.55 &0.862& 118.11 & 0.902\\
        $\gamma$-reweight & 154.03 &0.862& 140.72 &0.867& 150.46 &0.861& 116.91 & 0.906\\
        DiPmark($\alpha$=0.4) & 148.97 &0.863& 141.12 &0.860& 150.52 &0.864& 118.46 & 0.901\\
        DiPmark($\alpha$=0.3) & 153.32 &0.862& 139.00 &0.865& 150.04 &0.861& 114.95 & 0.905\\
        STA-1 & 155.15 & 0.864 & 132.26 & 0.873 & 151.03 & 0.870 & 117.96 & 0.906\\
        \midrule
        \methodname($h$=200) & 155.16 & 0.865 & 129.21 & 0.872& 138.28 &0.868 & 109.19 & 0.909\\
        \bottomrule
    \end{tabular}
    }
    \vspace{-0.3cm}
\end{table*}

\section{Experiments}

We implemented our pipeline in Python using the PyTorch framework and conducted experiments on four NVIDIA GeForce H100 GPUs. The experiments were performed on four datasets: Pickapic\_v2~\citep{kirstain2023pick}, MSCOCO~\citep{lin2014microsoft}, Flickr8k~\citep{hodosh2013framing}, and LAION~\citep{schuhmann2022laion}. We followed the generation configurations provided by each dataset, except that the generated images vary in size and aspect ratio. The auto-regressive image generation models for image generation and evaluation of the effectiveness of our proposed \methodname\ are Emu3~\citep{wang2024emu3} models.

% \paragraph{Baselines and parameters.} We evaluate the performance of our methods against various baselines, including two biased watermarking approaches, KGW~\citep{kirchenbauer2023watermark} and Unigram~\citep{zhao2023provable}, as well as five unbiased watermarking algorithms, i.e., ITS-edit~\citep{kuditipudi2023robust}, EXP-edit~\citep{kuditipudi2023robust}, $\gamma$-reweight~\citep{hu2023unbiased}, DiPmark~\citep{wu2023dipmark} and STA-1~\citep{mao2024watermark}.
% We select $\alpha \in\{ 0.3, 0.4\}$ for DiPmark, and $\delta \in \{0.5, 1.0, 1.5, 2.0\}$ and $\gamma=0.5$ for KGW watermark \citep{kirchenbauer2023watermark}, $\delta \in \{0.5, 1.0, 1.5, 2.0\}$ for Unigram~\citep{zhao2023provable}. For ITS-edit~\citep{kuditipudi2023robust}, EXP-edit~\citep{kuditipudi2023robust}, $\gamma$-reweight~\citep{hu2023unbiased} and STA-1~\citep{mao2024watermark}, we follow the settings in the original papers. We generate 500 examples for each tasks. We use the prefix 1-gram together with a secret key as the watermark keys.

\paragraph{Baselines and parameters.} We evaluate the performance of our methods by comparing them with multiple baseline approaches, which include two biased watermarking methods, i.e., KGW~\citep{kirchenbauer2023watermark} and Unigram~\citep{zhao2023provable}, as well as five unbiased watermarking algorithms, i.e., ITS-edit~\citep{kuditipudi2023robust}, EXP-edit~\citep{kuditipudi2023robust}, $\gamma$-reweight~\citep{hu2023unbiased}, DiPmark~\citep{wu2023dipmark}, and STA-1~\citep{mao2024watermark}. 
For DiPmark, we choose $\alpha \in \{0.3, 0.4\}$. For the KGW watermark~\citep{kirchenbauer2023watermark}, we use $\delta \in \{0.5, 1.0, 1.5, 2.0\}$ and $\gamma=0.5$. For Unigram~\citep{zhao2023provable}, we select $\delta \in \{0.5, 1.0, 1.5, 2.0\}$. The parameters for ITS-edit~\citep{kuditipudi2023robust}, EXP-edit~\citep{kuditipudi2023robust}, $\gamma$-reweight~\citep{hu2023unbiased}, and STA-1~\citep{mao2024watermark} are taken directly from their original papers. We generate 500 examples for each task. The watermark keys are constructed using the prefix 1-gram combined with a secret key.

% For ITS-edit~\citep{kuditipudi2023robust}, EXP-edit~\citep{kuditipudi2023robust}, $\gamma$-reweight~\citep{hu2023unbiased} and STA-1~\citep{mao2024watermark}, we follow the settings in the original papers. For \methodname\ we set the number of distribution channels $l=20$.

% \textbf{Settings.}
% \begin{table}
%   \caption{Sample table title}
%   \label{sample-table}
%   \centering
%   \begin{tabular}{lll}
%     \toprule
%     \multicolumn{2}{c}{Part}                   \\
%     \cmidrule(r){1-2}
%     Name     & Description     & Size ($\mu$m) \\
%     \midrule
%     Dendrite & Input terminal  & $\sim$100     \\
%     Axon     & Output terminal & $\sim$10      \\
%     Soma     & Cell body       & up to $10^6$  \\
%     \bottomrule
%   \end{tabular}
% \end{table}

\subsection{Detectability}
% \textbf
We evaluate the detectability of \methodname\ on the image generation task against baselines in Table~\ref{tab:main_res}. Following the evaluation metric of the previous works~\citep{kirchenbauer2023watermark,wu2023dipmark}, we report the True Positive Rate (TPR) at guaranteed False Positive Rates (FPR), i.e., TPR@FPR=$\{1\%, 0.1\%\}$. Notice that since the detectors for ITS-edit and EXP-edit lack a theoretical guarantee, we follow their original settings and report the true positive rate at their empirical false positive rate. Our method outperforms all other unbiased watermarks, showcasing superior detectability across different datasets.

\subsection{Robustness}
% We evaluate the robustness of the image watermark under various noise attacks in Table~\ref{tab:robustness}. We select $l_2$ noise with budgets \{0.25, 0.50, 1.00\} and $l_{inf}$ noise with budgets \{2/255, 4/255, 8/255\}. We test the detectability of different watermarks on generated images on Pickapic\_v2~\citep{kirstain2023pick} dataset and use TPR@FPR=$1$\% as the metric. \methodname\ outperforms all other unbiased watermarks when under attack.
We assess the robustness of the image watermark against various noise attacks, as presented in Table~\ref{tab:robustness_l2} and \ref{tab:robustness_linf}. Specifically, we apply $l_2$ noise with budgets \{0.25, 0.50, 1.00\} and $l_{\infty}$ noise with budgets \{2/255, 4/255, 8/255\} to simulate different attack scenarios.The detectability of each watermark is tested on images generated using the configuration of the Pickapic\_v2\citep{kirstain2023pick} dataset. We use True Positive Rate at a fixed False Positive Rate of 1\% (TPR@FPR=$1$\%) as the evaluation metric. The results demonstrate that \methodname\ consistently outperforms all other unbiased watermarking methods under these adversarial conditions, highlighting its superior robustness to noise attacks. This performance underscores its effectiveness in maintaining detectability even when subjected to significant perturbations.

\subsection{Image Quality}

% \paragraph{Qualitative Comparison.} We demonstrate some generation results in Table~\ref{tab:main_imqual}. Each pair of images are generated with the same settings including prompt and image size. The generated results with \methodname\ shows no obvious degradation in quality compared to those without watermark.

\paragraph{Qualitative Comparison} We present a qualitative comparison of generation results in Table~\ref{tab:main_imqual}. Each pair of images is generated under identical settings, including the prompt and image size. The results indicate that images generated with \methodname\ exhibit no noticeable degradation in visual quality compared to those produced without watermarking.

% \paragraph{Quantitative Evaluation.}
% For unbiasedness validation, we calculate the FID score~\citep{heusel2017gans} and CLIP score~\citep{radford2021learning} between the watermarked images and the unwatermarked images with the same prompts. The FID score measures the distance of feature vectors extracted using Inception models, and the CLIP
% score shows the cosine similarity between image embeddings. We use the scores between two sets of unwatermarked images as baseline. Results show that watermarked image using \methodname\ has similar quality as the unwatermarked ones.

\paragraph{Quantitative Evaluation} For unbiasedness validation, we compute the Fréchet Inception Distance (FID) score~\citep{heusel2017gans} and the CLIP score~\citep{radford2021learning} between watermarked and unwatermarked images generated from the same prompts. The FID score measures the distance between feature vectors extracted using Inception models, while the CLIP score quantifies the cosine similarity between image embeddings. We use the scores between two sets of unwatermarked images as baseline. The results demonstrate that images watermarked with \methodname\ maintain comparable quality to their unwatermarked counterparts, as evidenced by the similarity in FID and CLIP scores. This confirms that \methodname\ effectively preserves image quality while embedding a watermark.

% \subsection{Ablation study}

\vspace{-0.2cm}
\section{Conclusion}
\vspace{-0.2cm}
We introduce \methodname, a novel distortion-free watermarking method specifically designed for autoregressive image generation. By employing a clustering-based strategy that accounts for token similarity during retokenization, \methodname\ effectively mitigates sequence mismatches while preserving image fidelity. Experimental evaluations confirm that \methodname\ not only maintains the visual quality of generated images but also enhances watermark detectability, outperforming existing distortion-free methods. Our approach establishes a robust foundation for secure and trustworthy image synthesis, paving the way for more resilient watermarking mechanisms in future visual generation technologies.

\clearpage
\newpage
\bibliography{acl_latex}

\begin{thebibliography}{33}
\providecommand{\natexlab}[1]{#1}

\bibitem[{Aaronson(2022)}]{Aaronson2022}
Scott Aaronson. 2022.
\newblock \href {https://scottaaronson.blog/?p=6823} {My {AI} safety lecture for {UT} effective altruism,}.

\bibitem[{Brown(2020)}]{brown2020language}
Tom~B Brown. 2020.
\newblock Language models are few-shot learners.
\newblock \emph{arXiv preprint arXiv:2005.14165}.

\bibitem[{Christ et~al.(2023)Christ, Gunn, and Zamir}]{christ2023undetectable}
Miranda Christ, Sam Gunn, and Or~Zamir. 2023.
\newblock Undetectable watermarks for language models.
\newblock \emph{arXiv preprint arXiv:2306.09194}.

\bibitem[{Ci et~al.(2024)Ci, Yang, Song, and Shou}]{ci2024ringid}
Hai Ci, Pei Yang, Yiren Song, and Mike~Zheng Shou. 2024.
\newblock Ringid: Rethinking tree-ring watermarking for enhanced multi-key identification.
\newblock In \emph{European Conference on Computer Vision}, pages 338--354. Springer.

\bibitem[{Dathathri et~al.(2024)Dathathri, See, Ghaisas, Huang, McAdam, Welbl, Bachani, Kaskasoli, Stanforth, Matejovicova et~al.}]{dathathri2024scalable}
Sumanth Dathathri, Abigail See, Sumedh Ghaisas, Po-Sen Huang, Rob McAdam, Johannes Welbl, Vandana Bachani, Alex Kaskasoli, Robert Stanforth, Tatiana Matejovicova, and 1 others. 2024.
\newblock Scalable watermarking for identifying large language model outputs.
\newblock \emph{Nature}, 634(8035):818--823.

\bibitem[{Esser et~al.(2021)Esser, Rombach, and Ommer}]{esser2021taming}
Patrick Esser, Robin Rombach, and Bjorn Ommer. 2021.
\newblock Taming transformers for high-resolution image synthesis.
\newblock In \emph{Proceedings of the IEEE/CVF conference on computer vision and pattern recognition}, pages 12873--12883.

\bibitem[{Fernandez et~al.(2023)Fernandez, Couairon, J{\'e}gou, Douze, and Furon}]{fernandez2023stable}
Pierre Fernandez, Guillaume Couairon, Herv{\'e} J{\'e}gou, Matthijs Douze, and Teddy Furon. 2023.
\newblock The stable signature: Rooting watermarks in latent diffusion models.
\newblock In \emph{Proceedings of the IEEE/CVF International Conference on Computer Vision}, pages 22466--22477.

\bibitem[{Gunn et~al.(2024)Gunn, Zhao, and Song}]{gunn2024undetectable}
Sam Gunn, Xuandong Zhao, and Dawn Song. 2024.
\newblock An undetectable watermark for generative image models.
\newblock \emph{arXiv preprint arXiv:2410.07369}.

\bibitem[{Heusel et~al.(2017)Heusel, Ramsauer, Unterthiner, Nessler, and Hochreiter}]{heusel2017gans}
Martin Heusel, Hubert Ramsauer, Thomas Unterthiner, Bernhard Nessler, and Sepp Hochreiter. 2017.
\newblock Gans trained by a two time-scale update rule converge to a local nash equilibrium.
\newblock \emph{Advances in neural information processing systems}, 30.

\bibitem[{Hodosh et~al.(2013)Hodosh, Young, and Hockenmaier}]{hodosh2013framing}
Micah Hodosh, Peter Young, and Julia Hockenmaier. 2013.
\newblock Framing image description as a ranking task: Data, models and evaluation metrics.
\newblock \emph{Journal of Artificial Intelligence Research}, 47:853--899.

\bibitem[{Hu et~al.(2023)Hu, Chen, Wu, Wu, Zhang, and Huang}]{hu2023unbiased}
Zhengmian Hu, Lichang Chen, Xidong Wu, Yihan Wu, Hongyang Zhang, and Heng Huang. 2023.
\newblock Unbiased watermark for large language models.
\newblock \emph{arXiv preprint arXiv:2310.10669}.

\bibitem[{Kirchenbauer et~al.(2023)Kirchenbauer, Geiping, Wen, Katz, Miers, and Goldstein}]{kirchenbauer2023watermark}
John Kirchenbauer, Jonas Geiping, Yuxin Wen, Jonathan Katz, Ian Miers, and Tom Goldstein. 2023.
\newblock A watermark for large language models.
\newblock \emph{arXiv preprint arXiv:2301.10226}.

\bibitem[{Kirstain et~al.(2023)Kirstain, Polyak, Singer, Matiana, Penna, and Levy}]{kirstain2023pick}
Yuval Kirstain, Adam Polyak, Uriel Singer, Shahbuland Matiana, Joe Penna, and Omer Levy. 2023.
\newblock Pick-a-pic: An open dataset of user preferences for text-to-image generation.
\newblock \emph{Advances in Neural Information Processing Systems}, 36:36652--36663.

\bibitem[{Kuditipudi et~al.(2023)Kuditipudi, Thickstun, Hashimoto, and Liang}]{kuditipudi2023robust}
Rohith Kuditipudi, John Thickstun, Tatsunori Hashimoto, and Percy Liang. 2023.
\newblock Robust distortion-free watermarks for language models.
\newblock \emph{arXiv preprint arXiv:2307.15593}.

\bibitem[{Lin et~al.(2014)Lin, Maire, Belongie, Hays, Perona, Ramanan, Doll{\'a}r, and Zitnick}]{lin2014microsoft}
Tsung-Yi Lin, Michael Maire, Serge Belongie, James Hays, Pietro Perona, Deva Ramanan, Piotr Doll{\'a}r, and C~Lawrence Zitnick. 2014.
\newblock Microsoft coco: Common objects in context.
\newblock In \emph{Computer vision--ECCV 2014: 13th European conference, zurich, Switzerland, September 6-12, 2014, proceedings, part v 13}, pages 740--755. Springer.

\bibitem[{Liu et~al.(2023)Liu, Li, Wu, and Lee}]{liu2023visual}
Haotian Liu, Chunyuan Li, Qingyang Wu, and Yong~Jae Lee. 2023.
\newblock Visual instruction tuning.
\newblock \emph{Advances in neural information processing systems}, 36:34892--34916.

\bibitem[{Mao et~al.(2024)Mao, Wei, Chen, Fang, and Chau}]{mao2024watermark}
Minjia Mao, Dongjun Wei, Zeyu Chen, Xiao Fang, and Michael Chau. 2024.
\newblock A watermark for low-entropy and unbiased generation in large language models.
\newblock \emph{arXiv preprint arXiv:2405.14604}.

\bibitem[{Radford et~al.(2021)Radford, Kim, Hallacy, Ramesh, Goh, Agarwal, Sastry, Askell, Mishkin, Clark et~al.}]{radford2021learning}
Alec Radford, Jong~Wook Kim, Chris Hallacy, Aditya Ramesh, Gabriel Goh, Sandhini Agarwal, Girish Sastry, Amanda Askell, Pamela Mishkin, Jack Clark, and 1 others. 2021.
\newblock Learning transferable visual models from natural language supervision.
\newblock In \emph{International conference on machine learning}, pages 8748--8763. PmLR.

\bibitem[{Raffel et~al.(2020)Raffel, Shazeer, Roberts, Lee, Narang, Matena, Zhou, Li, and Liu}]{raffel2020exploring}
Colin Raffel, Noam Shazeer, Adam Roberts, Katherine Lee, Sharan Narang, Michael Matena, Yanqi Zhou, Wei Li, and Peter~J Liu. 2020.
\newblock Exploring the limits of transfer learning with a unified text-to-text transformer.
\newblock \emph{Journal of machine learning research}, 21(140):1--67.

\bibitem[{Rombach et~al.(2022)Rombach, Blattmann, Lorenz, Esser, and Ommer}]{rombach2022high}
Robin Rombach, Andreas Blattmann, Dominik Lorenz, Patrick Esser, and Bj{\"o}rn Ommer. 2022.
\newblock High-resolution image synthesis with latent diffusion models.
\newblock In \emph{Proceedings of the IEEE/CVF conference on computer vision and pattern recognition}, pages 10684--10695.

\bibitem[{Schuhmann et~al.(2022)Schuhmann, Beaumont, Vencu, Gordon, Wightman, Cherti, Coombes, Katta, Mullis, Wortsman et~al.}]{schuhmann2022laion}
Christoph Schuhmann, Romain Beaumont, Richard Vencu, Cade Gordon, Ross Wightman, Mehdi Cherti, Theo Coombes, Aarush Katta, Clayton Mullis, Mitchell Wortsman, and 1 others. 2022.
\newblock Laion-5b: An open large-scale dataset for training next generation image-text models.
\newblock \emph{Advances in neural information processing systems}, 35:25278--25294.

\bibitem[{Song et~al.(2020)Song, Meng, and Ermon}]{song2020denoising}
Jiaming Song, Chenlin Meng, and Stefano Ermon. 2020.
\newblock Denoising diffusion implicit models.
\newblock \emph{arXiv preprint arXiv:2010.02502}.

\bibitem[{Sun et~al.(2023)Sun, Yu, Cui, Zhang, Zhang, Wang, Gao, Liu, Huang, and Wang}]{sun2023emu}
Quan Sun, Qiying Yu, Yufeng Cui, Fan Zhang, Xiaosong Zhang, Yueze Wang, Hongcheng Gao, Jingjing Liu, Tiejun Huang, and Xinlong Wang. 2023.
\newblock Emu: Generative pretraining in multimodality.
\newblock \emph{arXiv preprint arXiv:2307.05222}.

\bibitem[{Team(2024)}]{team2024chameleon}
Chameleon Team. 2024.
\newblock Chameleon: Mixed-modal early-fusion foundation models.
\newblock \emph{arXiv preprint arXiv:2405.09818}.

\bibitem[{Vaswani et~al.(2017)Vaswani, Shazeer, Parmar, Uszkoreit, Jones, Gomez, Kaiser, and Polosukhin}]{vaswani2017attention}
Ashish Vaswani, Noam Shazeer, Niki Parmar, Jakob Uszkoreit, Llion Jones, Aidan~N Gomez, Lukasz Kaiser, and Illia Polosukhin. 2017.
\newblock Attention is all you need.
\newblock \emph{Advances in neural information processing systems}, 30.

\bibitem[{Wang et~al.(2024)Wang, Zhang, Luo, Sun, Cui, Wang, Zhang, Wang, Li, Yu et~al.}]{wang2024emu3}
Xinlong Wang, Xiaosong Zhang, Zhengxiong Luo, Quan Sun, Yufeng Cui, Jinsheng Wang, Fan Zhang, Yueze Wang, Zhen Li, Qiying Yu, and 1 others. 2024.
\newblock Emu3: Next-token prediction is all you need.
\newblock \emph{arXiv preprint arXiv:2409.18869}.

\bibitem[{Wang et~al.(2023)Wang, He, Li, Li, Yu, Ma, Li, Chen, Chen, Wang et~al.}]{wang2023internvid}
Yi~Wang, Yinan He, Yizhuo Li, Kunchang Li, Jiashuo Yu, Xin Ma, Xinhao Li, Guo Chen, Xinyuan Chen, Yaohui Wang, and 1 others. 2023.
\newblock Internvid: A large-scale video-text dataset for multimodal understanding and generation.
\newblock \emph{arXiv preprint arXiv:2307.06942}.

\bibitem[{Wen et~al.(2023)Wen, Kirchenbauer, Geiping, and Goldstein}]{wen2023tree}
Yuxin Wen, John Kirchenbauer, Jonas Geiping, and Tom Goldstein. 2023.
\newblock Tree-ring watermarks: Fingerprints for diffusion images that are invisible and robust.
\newblock \emph{arXiv preprint arXiv:2305.20030}.

\bibitem[{Wu et~al.(2023)Wu, Hu, Zhang, and Huang}]{wu2023dipmark}
Yihan Wu, Zhengmian Hu, Hongyang Zhang, and Heng Huang. 2023.
\newblock Dipmark: A stealthy, efficient and resilient watermark for large language models.
\newblock \emph{arXiv preprint arXiv:2310.07710}.

\bibitem[{Yang et~al.(2024)Yang, Zeng, Chen, Fang, Zhang, and Yu}]{yang2024gaussian}
Zijin Yang, Kai Zeng, Kejiang Chen, Han Fang, Weiming Zhang, and Nenghai Yu. 2024.
\newblock Gaussian shading: Provable performance-lossless image watermarking for diffusion models.
\newblock In \emph{Proceedings of the IEEE/CVF Conference on Computer Vision and Pattern Recognition}, pages 12162--12171.

\bibitem[{Zhang et~al.(2018)Zhang, Isola, Efros, Shechtman, and Wang}]{zhang2018unreasonable}
Richard Zhang, Phillip Isola, Alexei~A Efros, Eli Shechtman, and Oliver Wang. 2018.
\newblock The unreasonable effectiveness of deep features as a perceptual metric.
\newblock In \emph{Proceedings of the IEEE conference on computer vision and pattern recognition}, pages 586--595.

\bibitem[{Zhao et~al.(2023)Zhao, Ananth, Li, and Wang}]{zhao2023provable}
Xuandong Zhao, Prabhanjan Ananth, Lei Li, and Yu-Xiang Wang. 2023.
\newblock Provable robust watermarking for ai-generated text.
\newblock \emph{arXiv preprint arXiv:2306.17439}.

\bibitem[{Zheng et~al.(2022)Zheng, Vuong, Cai, and Phung}]{zheng2022movq}
Chuanxia Zheng, Tung-Long Vuong, Jianfei Cai, and Dinh Phung. 2022.
\newblock Movq: Modulating quantized vectors for high-fidelity image generation.
\newblock \emph{Advances in Neural Information Processing Systems}, 35:23412--23425.

\end{thebibliography}

\appendix
\onecolumn
\section{Missing Proofs}~\label{sec:missing proof}
\begin{theorem}
    Cluster-based reweight is distortion-free.
\end{theorem}
\begin{proof}
    We will show for all $x$, $\mathbb{E}_{\theta}[P_W(x|\x_{1:n},\theta)]=P_M(x|\x_{1:n})$. Assuming w.l.o.g. $x\in c_1$, the probability of sampling $x$ comes from two parts: a) $x\in c_{i'(\theta)}$ and b) $x\in c_{i''(\theta)}$. 
    
    \textbf{Case 1.} If $h\Pr(c_1)\leq1$, as $c_{i'(\theta)}$ is uniformly selected from the cluster set, $\Pr(c_{i'(\theta)}=c_1)=1/h$. Besides, the probability of sampling $c_1$ from the overflow distribution $\cP_c$ is $0$. In this case, 
    \begin{equation}
    \begin{split}
    \mathbb{E}_{\theta}[P_W(x|\x_{1:n},\theta)]&=\Pr(c_{i'(\theta)}=c_1)\Pr(j<h\Pr(c_1))\frac{P_M(x|\x_{1:n})}{\Pr(c_{1})}\\
    &=\frac{1}{h}*h\Pr(c_1)*\frac{P_M(x|\x_{1:n})}{\Pr(c_{1})}\\
    &=P_M(x|\x_{1:n}).
    \end{split}
    \end{equation}
    \textbf{Case 2.} If $h\Pr(c_1)>1$, we still have $\Pr(c_{i'(\theta)}=c_1)=1/h$ and the probability of sampling $c_1$ from the overflow distribution $\cP_c$ is $\Pr_{c\sim\cP_c}(c=c_1):=\frac{\min\{0,h\Pr(c_1)-1\}}{\sum_{i=1}^h\min\{0,h\Pr(c_i)-1\}}$. In this case
    \begin{equation}
    \begin{split}
    \mathbb{E}_{\theta}[P_W(x|\x_{1:n},\theta)]&=\frac{1}{h}\Pr(j<h\Pr(c_1))\frac{P_M(x|\x_{1:n})}{\Pr(c_{1})}+\frac{1}{h}\sum_{i=2}^h\Pr(j>h\Pr(c_i))\Pr_{c\sim\cP_c}(c=c_1)\frac{P_M(x|\x_{1:n})}{\Pr(c_{1})}\\
    &=\frac{1}{h}\frac{P_M(x|\x_{1:n})}{\Pr(c_{1})}+\frac{1}{h}\Pr_{c\sim\cP_c}(c=c_1)\frac{P_M(x|\x_{1:n})}{\Pr(c_{1})}\sum_{i=2}^h\Pr(j>h\Pr(c_i))\\
    &=\frac{1}{h}\frac{P_M(x|\x_{1:n})}{\Pr(c_{1})}+\frac{1}{h}\Pr_{c\sim\cP_c}(c=c_1)\frac{P_M(x|\x_{1:n})}{\Pr(c_{1})}\sum_{i=1}^h\min\{0,1-h\Pr(c_i)\}.
    \end{split}
    \end{equation}
    As $\sum_{i=1}^h\Pr(c_i)=1$, $\sum_{i=1}^h\min\{0,1-h\Pr(c_i)\}=\sum_{i=1}^h\min\{0,h\Pr(c_i)-1\}$, we have
    \begin{equation}
    \begin{split}
    \mathbb{E}_{\theta}[P_W(x|\x_{1:n},\theta)]&=\frac{1}{h}\frac{P_M(x|\x_{1:n})}{\Pr(c_{1})}+\frac{1}{h}\Pr_{c\sim\cP_c}(c=c_1)\frac{P_M(x|\x_{1:n})}{\Pr(c_{1})}\sum_{i=1}^h\min\{0,1-h\Pr(c_i)\},\\
    &=\frac{1}{h}\frac{P_M(x|\x_{1:n})}{\Pr(c_{1})}+\frac{1}{h}\frac{P_M(x|\x_{1:n})}{\Pr(c_{1})}\frac{\min\{0,h\Pr(c_1)-1\}}{\sum_{i=1}^h\min\{0,h\Pr(c_i)-1\}}\sum_{i=1}^h\min\{0,1-h\Pr(c_i)\},\\
    &=\frac{1}{h}\frac{P_M(x|\x_{1:n})}{\Pr(c_{1})}+\frac{1}{h}\frac{P_M(x|\x_{1:n})}{\Pr(c_{1})}(h\Pr(c_1)-1),\\
    &=P_M(x|\x_{1:n}).
    \end{split}
    \end{equation}
    Combining Case 1 and 2, we can conclude that cluster-based reweight is distortion-free.
\end{proof}
\end{document}